# Khmer Spellchecking: A Holistic Approach

Rina Buoy[†]     Sovisal Chenda[†]     Nguonly Taing[†]     Marry Kong[†]

[†]Techo Startup Center, Cambodia

rina.buoy@techostartup.center

**Abstract**

Compared to English and other high-resource languages, spellchecking for Khmer remains an unresolved problem due to several challenges. First, there are misalignments between words in the lexicon (e.g., មើល, ឃើញ: separate words) and the word segmentation model (e.g., មើលឃើញ: single word). Second, a Khmer word can be written in different forms (i.e., stacked vs. unstacked; e.g., តំលៃ vs. តម្លៃ). Third, Khmer compound words are often loosely and easily formed, and these compound words are not always found in the lexicon (e.g., ភាពរឹកចម្រើន). Fourth, some proper nouns (e.g., ប៉ុនឡេង) may be flagged as misspellings due to the absence of a Khmer named-entity recognition (NER) model. Unfortunately, existing solutions do not adequately address these challenges. This paper proposes a holistic approach to the Khmer spellchecking problem by integrating Khmer subword segmentation, Khmer NER, Khmer grapheme-to-phoneme (G2P) conversion, and a Khmer language model to tackle these challenges, identify potential correction candidates, and rank the most suitable candidate. Experimental results show that the proposed approach achieves a state-of-the-art Khmer spellchecking accuracy of up to 94.4%, compared to existing solutions. The benchmark datasets for Khmer spellchecking and NER tasks in this study will be made publicly available.

## 1 Introduction

Spellchecking is one of the fundamental natural language processing tasks with a wide range of practical applications. While spellchecking for English and other high-resource languages is relatively straightforward, spellchecking for Khmer remains a challenging, unsolved problem. First, the definition of words is relatively ambiguous in Khmer [1; 2; 3]. As a result, a Khmer word segmentation model does not necessarily segment words as they appear in a lexicon. For example, the segmentation model may treat the compound word អ្នកចម្រៀង (អ្នក+ចម្រៀង; meaning *singer*) as a single unit or word. However, this single unit does not appear in the official lexicon[1] and will be flagged as misspelled. Second, some Khmer words can be written in either stacked or unstacked forms [1; 3]. For example, the word ចំរៀង (meaning to *sing*) can also be written as ចម្រៀង; both forms have the same pronunciation. However, only the later appears in the official lexicon. Third, Khmer compound words are often easily and loosely constructed. For instance, the prefix អ្នក can be added to any action verb to denote the doer of the action. Since the official lexicon does not contain all possible realizations of compounds starting with the prefix អ្នក, some of these compounds will be flagged as misspelled. Fourth, due to the lack of Khmer named-entity recognition (NER), some proper nouns (e.g., ប៉ុនឡេង) will be flagged as misspelled, as they do not usually appear in the official lexicon.

However, existing Khmer spellchecking solutions are primarily based on basic lexical search. Specifically, for a given input sentence, word segmentation or tokenization is performed to obtain individual words. For each misspelled word, edit distances are computed against all entries in a given lexicon. The entry with the lowest distance is often returned as the most probable correction [4]. Such basic lexical search methods are insufficient to address the challenges mentioned above.

---

[1]https://nckl.rac.gov.kh/

Thus, in this paper, we present a holistic approach for Khmer spellchecking. Specifically, the proposed approach utilizes Khmer word segmentation at the sub-word level, making it possible to handle complex Khmer compound words. Additionally, the proposed approach incorporates Khmer grapheme-to-phoneme (G2P) conversion to easily locate words with different spellings but the same pronunciation in a given lexicon [4]. Furthermore, the approach integrates a Khmer NER model to address proper names that do not appear in the lexicon. Lastly, instead of greedily suggesting the correction with the lowest edit distance, the proposed approach uses a Khmer language model to rank and optimally identify the best corrections.

Our contributions can be summarized as follows:

1. We propose the first holistic approach to Khmer spellchecking, addressing multiple challenges associated with the Khmer language.

2. We are the first to incorporate a Khmer language model to introduce contextual information into Khmer spellchecking.

3. Our proposed approach achieves state-of-the-art performance in Khmer spellchecking.

4. We contribute the first open benchmark datasets for Khmer spellchecking and NER tasks.

## 2 Related Work

Although there are not many published solutions in the area of Khmer spellchecking, existing solutions can be grouped into word-level and syllable-level approaches.

The syllable-level methods [5; 6] split a given input sentence into a sequence of character clusters or syllables. Individual clusters or syllables are then corrected before being combined to form valid words. However, such simplistic methods can fail to address the challenges of Khmer spellchecking and do not always identify correct words. For instance, the word ទស្សន:ទាន is a misspelling of the correct word ទស្សនទាន (meaning *concept*). Although at the syllable level, both syllables (i.e., ទស្សន: and ទាន) are individually correct, this approach fails to detect the error. The same issue arises with អត្តសញ្ញាជាតិ (misspelled) vs. អត្តសញ្ញាណជាតិ (correct; meaning *national identity*).

Many word-level methods, such as NextSpell [2], SSBIC [3], and KhmerLang [4], are similar to a certain extent. Typically, these methods first segment an input sentence into a sequence of words. Individual misspelled words are then corrected. The correction process starts by computing edit distances between a misspelled word and all words in a given lexicon, returning the possible candidates with a distance less than a predefined threshold [4]. Many of these solutions primarily work on the grapheme level rather than the phoneme level. Additionally, these solutions do not take context into account when suggesting corrections. Thus, existing word-level methods can fail to correct complex Khmer out-of-vocabulary compound words (e.g., អ្នកដឹក; meaning *alcoholic*), identify different spelling forms (e.g., អ្នកចម្រៀង vs. អ្នកចំរៀង; meaning *singer*), and utilize contextual information effectively. Instead of using the same official lexicon, these methods rely on their own built lexicons.

## 3 Proposed Solution

Before explaining our proposed holistic approach for Khmer spellchecking, we first introduce the key components of our approach. Specifically, we discuss Khmer word segmentation, Khmer grapheme-to-phoneme (G2P) conversion, Khmer named-entity recognition (NER), and Khmer language modeling.

### 3.1 Khmer Word Segmentation

To identify misspelled words, we need to locate word boundaries in a given sentence. Since Khmer does not use visible word delimiters, word segmentation is performed first [1]. Word segmentation errors can also result in misspelled words. For instance, the word មរតក may be erroneously segmented into មរ and តក, neither of which are correct words. To mitigate this issue, a word segmentation model must be invariant to spelling mistakes, which can be achieved through data augmentation by introducing random spelling errors during training time. For example, the character ឡ can be removed from the word សាលារៀន (meaning *school*).

---
[2]https://nextspell.com/
[3]https://sbbic.org/
[4]https://khmerlang.com/

We trained a word segmentation model based on the Llama 2 architecture [7] using llama2.c[5] instead of bi-directional maximum matching (BMM)[8], conditional random field [2], and recurrent neural networks (RNN)[9] to enhance textual information through unmasked self-attention layers. The detailed specifications of the Llama-based Khmer word segmentation model are provided in Table 1. The input dimension corresponds to the unique number of Khmer characters and other symbols (e.g., ក, ខ, etc.). The model returns four labels (i.e., output dimension): *no_space* for no word boundaries, *space* for word boundaries, _ for compound words (e.g., ខោ_អាវ; meaning *clothes*), ˜ for prefixes (e.g., អ្នក˜ចម្រៀង; meaning *singer*), and ˆ for suffixes (e.g., ទំនើបˆកម្ម; meaning *modernization*).

The original Llama architecture [7] is designed for autoregressive language modeling. However, it can be adapted for Khmer word segmentation, which is essentially a sequence labeling/tagging task. This is achieved by replacing causal attention masks with full attention masks. In other words, to predict a tag or label for a given character, the model can attend to all input characters, as illustrated in Figure 1.

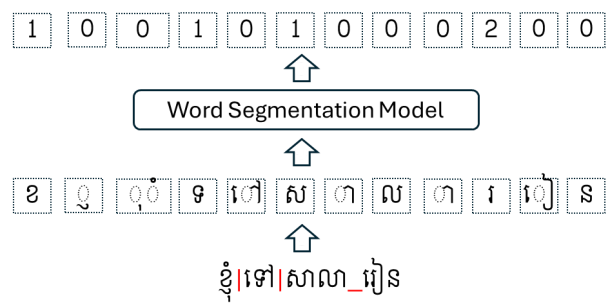

Figure 1. An illustration of Khmer word segmentation. |: word boundary notation (label: 1). _: compound word notation (label: 2). 0: no space between characters.

---
[5]https://github.com/karpathy/llama2.c

For training data and model training, we used the existing word segmentation solution [2] to generate the training data (i.e., segmented sentences). We manually curated the resulting segmented sentences and corrected any segmentation errors. The curated segmented sentences were further augmented during training by randomly deleting (e.g., កណ្តាល to កណ្តា), substituting (e.g., កណ្តាល to កន្តាល), or adding characters (e.g., កណ្តាល to កណ្តាលរ) to simulate spelling errors. This data augmentation helps the model become invariant to spelling errors. We evaluated the model on the segmented closed-test set [10]. The average precision, recall, and F1-score across all tags are 0.99, 0.88, and 0.94, respectively.

Table 1. Configurations of the Khmer word segmentation model.

| Parameter | Value |
|---|---|
| Model dimension | 192 |
| Number of layers | 6 |
| Attention heads | 8 |
| Feedforward dimension | 3 |
| Input dimension | 156 |
| Output dimension | 4 |
| Max Sequence Length | 512 |
| Dropout | 0.1 |

### 3.2 Khmer G2P Model

To identify different spelling variations of a given Khmer word (e.g., ចំរៀង vs. ចម្រៀង), we cannot rely on edit distance computed on graphemes alone. Since different spelling variations have roughly the same pronunciation, the edit distance computed on phonemes should be close to or equal to zero. To convert graphemes to phonemes, we trained a Khmer G2P model using the publicly available dataset provided by Google[6] [11]. The dataset was split into training and test sets using a 90%-10% rule.

We used the standard Transformer-based encoder-decoder architecture [12], originally developed for machine translation, to convert a Khmer word into its phoneme representation. The encoder takes a sequence of Khmer characters as input and returns hidden representations, which are passed to the decoder to generate a sequence of phonemes autoregressively. The detailed configurations of the encoder and decoder are provided in Tables 2 and 3. The input and output dimensions of the encoder and decoder are 167, representing the total number of unique Khmer characters (e.g., ក, ខ, etc.) and phonemes (e.g., a, aa, etc.), as illustrated in Figure 2. The resulting character error rate (CER) on the test set is 6.6%. The CER of the weighted finite state transducer (WFST) technique [13] computed on the same test set is approximately 7.1%.

---
[6]https://github.com/google/language-resources

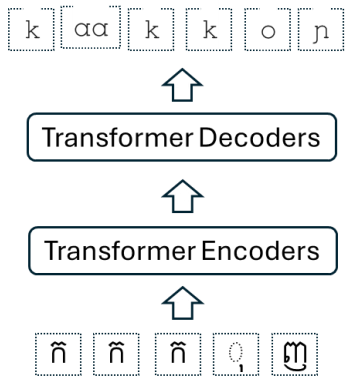

Figure 2. An illustration of character-level encoder-decoder Khmer G2P, mapping a Khmer word to its phoneme representation.

Table 2. Encoder configurations of the Khmer G2P model.

| Parameter | Value |
| --- | --- |
| Model dimension | 192 |
| Number of layers | 3 |
| Attention heads | 8 |
| Feedforward dimension | 4 |
| Input dimension | 167 |
| Max Sequence Length | 100 |
| Dropout | 0.1 |

### 3.3 Khmer NER Model

Proper nouns, such as names of people, places, entities, etc., should not be spellchecked. To identify whether words are names, we need a Khmer named-entity recognition (NER) model. NER is a span labeling task that identifies a span of words belonging to a particular named entity [14]. We constructed a Khmer NER dataset and trained a Khmer model accordingly. Since no Khmer NER datasets are available, we used Google Translate[7] to translate publicly available English-based NER datasets into Khmer with manual curation. This simplistic approach is not ideal, but manual acquisition of Khmer NER data is a time-consuming and labor-intensive task, but can yield high quality training data.

Table 3. Decoder configurations of the Khmer G2P model.

| Parameter | Value |
| --- | --- |
| Model dimension | 192 |
| Number of layers | 3 |
| Attention heads | 8 |
| Feedforward dimension | 4 |
| Output dimension | 167 |
| Max Sequence Length | 100 |
| Dropout | 0.1 |

The Khmer NER model has the same architecture as the Khmer word segmentation model described in Section 3.1, except that the output dimension is two, indicating whether a character is part of a named entity or not, as illustrated in Figure 3. Previous work on the Khmer NER task [15] was based on the classical conditional random field technique. The recognition accuracy of the trained model is approximately 92.5% on the test dataset.

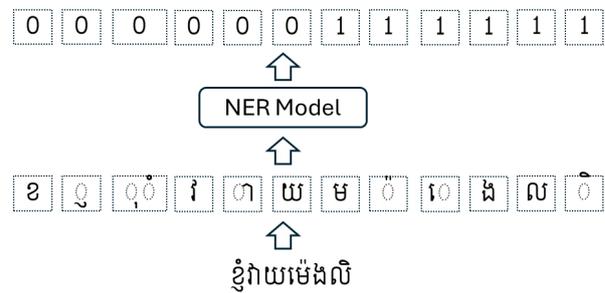

Figure 3. An illustration of character-level Khmer named-entity recognition. 1: named entity. 0: not named entity.

---

[7]https://translate.google.com/

## 3.4 Khmer Language Model

Given a misspelled word, there may be multiple possible correction candidates. To identify the most probable one, contextual information is needed. A Khmer language model can be used for this purpose.

We trained a Llama-based Khmer language model [7], the specifications of which are provided in Table 4. The training data comprise various sources, including news articles and Wikipedia pages, totaling approximately 300K articles. No word segmentation or minor preprocessing steps were performed on the training data. The model used character-level tokenization instead of subword tokenization schemes, such as sentencepiece [16] or byte-pair encoding [17], as illustrated in Figure 4.

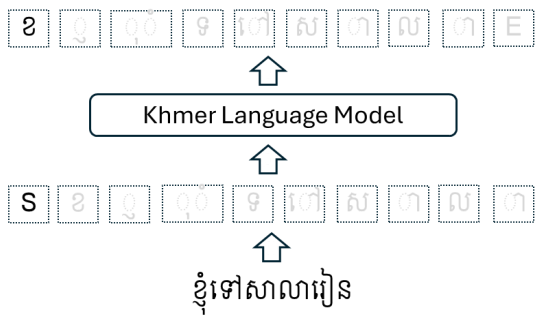

Figure 4. An illustration of Khmer decoder-only language model. An $S$ token (start of a sentence) is used to predict the first character.

Table 4. Configurations of the Khmer language model.

| Parameter | Value |
| --- | --- |
| Model dimension | 384 |
| Number of layers | 8 |
| Attention heads | 8 |
| Feedforward dimension | 3 |
| Input dimension | 2,892 |
| Output dimension | 2,892 |
| Max Sequence Length | 6,144 |
| Dropout | 0.0 |

## 3.5 Holistic Khmer Spellchecking Solution

The integrated holistic workflow for Khmer spellchecking is provided in Figure 5. The processes are as follows:

1. We use a trained Khmer word segmentation model to segment an input sentence into a sequence of words. At the same time, we also run a trained NER model on the input sentence to extract any names.

2. For each non-NER misspelled word, we identify the closest corrections by computing edit distances in both phonemes and graphemes against entries in a given lexicon. Then, we return the top $k$ correction candidates for each misspelled word that have an edit distance less than $\epsilon$ [4]. For a misspelled compound word, we repeat this step for individual subwords as well. For instance, if សាលា_រៀង is a misspelled version of សាលា_រៀន, we run this step for សាលារៀង (compound), សាលា (subword), and រៀង (subword). The corrections at the subword level are then combined to obtain the compound-level corrections.

3. We use a trained Khmer language model to identify the best correction for each misspelled word by maximizing sentence-level likelihood.

Mathethically, step 3 can be expressed as:

$$\hat{H} = \arg\max_{H} p(H) \quad (1)$$

where $H$ is a possible correction hypothesis for a given input sentence. $p(H)$ is a likelihood of $H$ and given by a trained Khmer language model. $\hat{H}$ is the most likely correction hypothesis.

## 4 Experiments

### 4.1 Datasets

Since there are no standard datasets for evaluating the accuracy of a Khmer spellchecker, we synthetically generated a benchmark dataset. We started with 321 commonly misspelled words, some samples of which are listed in Table 5, and used Claude-3.5-sonnet

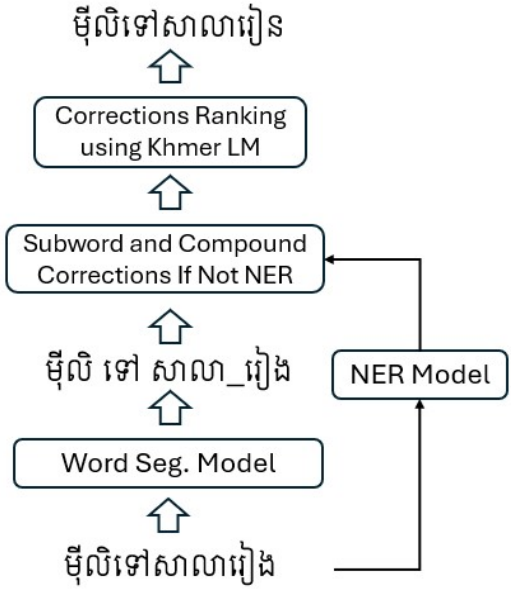

Figure 5. The proposed holistic Khmer spellchecking approach. Seg.: Segmentation. NER: Named-entity recognition. LM: Language model.

Table 5. Some samples of correct vs. misspelled words.

| Correct | Misspelled |
|---------|------------|
| សមាគម | សមាគុមន៍ |
| ទន់ភ្លន់ | ទន់ភ្លូន |
| កុមារភាព | ក្មមារភាព |
| កាយប្បទិ្ឋ | កាយវិទ្ឋិ |
| បំផុះ | បំផុះ |
| តន្ត្រីករ | តន្ត្រីករ |
| រោគសញ្ញា | រោកសញ្ញា |
| នគរូបនីយកម្ម | ណាគរូបនីយកម្ម |
| ជ្រក់ជ្រេញ | ជ្រកជ្រេញ |

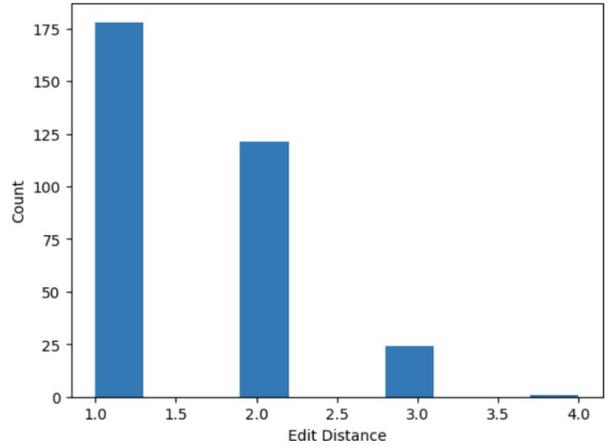

Figure 6. Edit distance distribution of the misspelled in Dataset A.

(version: 20240620) [18] to create short sentences, each containing a misspelled word. Claude-3.5-sonnet appears to be the best-performing large language model (LLM) for the Khmer language at the time of writing, compared with ChatGPT-4.0 [19] and other open-source LLMs, such as Llama-3.1 [20]. The edit distance distribution is provided in Figure 6. We call this 'Dataset A'. Some examples of the generated sentences are provided in Table 6.

Additionally, since a Khmer spellchecker should not flag a named entity as a misspelled word, we used 108 common person names[8] in Khmer and Claude-3.5-sonnet (version: 20240620) [18] to generate short sentences for evaluating the proposed approach. This named-entity dataset can be used to evaluate both spellchecking and Khmer NER performance. We call this 'Dataset B'. Some examples of the person names and generated sentences are provided in Table 7.

Both Datasets A and B will be made available to the community, along with this manuscript.

---
[8]https://github.com/silnrsi/khmerlbdict

### 4.2 Evaluation Metrics

We use an accuracy metric to measure spellchecking performance. Accuracy allows us to determine the percentage of misspelled words that are correctly identified and corrected.

Since there can be multiple correction hypotheses and these hypotheses are ranked by a Khmer language model, we compute the accuracy metrics at different top $n$ levels. For example, the accuracy at the top three (i.e., Top $3$) is the percentage of misspelled words that are correctly identified and corrected if the true correction is within the top three hypotheses.

Table 6. Some examples of LLM-generated sentences with the seeding misspelled words in Dataset A.

| Misspelled sentence |
| --- |
| ខ្ញុំចូលរួមក្នុង<u>សមាគមន៍</u>មួយដែលជួយដល់កុមារកំព្រា។ |
| ខ្ញុំនឹកឃើញពី<u>កូ</u>មារភាពរបស់ខ្ញុំជានិច្ច។ |
| ពួកគេបានបំផ្ទុះ<u>គ្រាប់បែក</u>នៅក្នុងអគារនោះ។ |
| គ្រូពេទ្យបានពិនិត្យ<u>រោគសញ្ញា</u>របស់អ្នកជំងឺយ៉ាង... |

Table 7. Some examples of LLM-generated sentences with the seeded person names in Dataset B.

| Name | Sentence |
| --- | --- |
| កន្និដ្ឋា | ខ្ញុំបានជួប លោកស្រី <u>កន្និដ្ឋា</u> នៅផ្សារ។ |
| ខៀវវិទ្ធ | លោក <u>ខៀវវិទ្ធ</u> គឺជាគ្រូបង្រៀនដ៏ល្អម្នាក់។ |
| ខេរីសា | ថ្ងៃនេះ លោកស្រី <u>ខេរីសា</u> នឹងទៅផ្សារ។ |

## 5 Results And Discussion

We used the official Khmer dictionary (version 2022)[9] released by the Royal Academy of Cambodia as the lexicon in the subsequent experiments. Any words spelled differently from this official lexicon are treated as misspelled. However, the proposed approach can work with any provided lexicons. For each misspelled word, we search for the top three correction candidates (i.e., $k = 3$) in the lexicon, which have a maximum edit distance of three ($\epsilon = 3$).

The sentences generated by Claude-3.5-sonnet are not guaranteed to be free from spelling errors, as Claude-3.5-sonnet was trained on large-scale data, which may not always align with the official lexicon. To ensure proper performance evaluation, we only perform spellchecking on the seeded misspelled words (i.e., underlined words in Table 6).

### 5.1 Results

We begin by providing the spellchecking results on Dataset A with different numbers of top hypotheses (i.e., $n$). At Top 3 (i.e., T3), we return the top three most likely correction hypotheses, ranked by the Khmer language model. As long as one of these three hypotheses contains the true word, we consider the spellcheck results to be correct.

As shown in Table 8, the total accuracy increases significantly as we increase the number of top correction hypotheses. Specifically, the total accuracy rises from 80.4% at Top 1 to 87.5% at Top 2, and to 91.9% at Top 3. This highlights that the Khmer language model used in the proposed approach can retain only the most relevant correction hypotheses. The table also indicates that the correction accuracy correlates negatively with edit distance, except at an edit distance of 4, where there are not enough samples. This means that the more a misspelled word deviates from its true form, the harder it is to correct.

On Dataset B, which contains person names not found in the lexicon, the proposed approach should not attempt to correct these names. Table 9 shows that we can correctly identify and refrain from correcting person names 107 out of 108 times. This corresponds to an accuracy of approximately 99.1%.

Table 8. Spellchecking results on Dataset A. T$k$: Top $k$. Corr.: Correct. Acc.: Accuracy in %. **Bold**: best.

| Case | Edit Distance | | | | |
| --- | --- | --- | --- | --- | --- |
| | 1 | 2 | 3 | 4 | Total |
| Total | 177 | 119 | 24 | 1 | 321 |
| Corr.@T1 | 143 | 96 | 18 | 1 | 258 |
| Acc.@T1 | 80.8 | 80.7 | 75.0 | 100.0 | 80.4 |
| Corr.@T2 | 156 | 106 | 18 | 1 | 281 |
| Acc.@T2 | 88.1 | 89.1 | 75.0 | 100.0 | 87.5 |
| Corr.@T3 | 167 | 109 | 18 | 1 | 295 |
| Acc.@T3 | **94.4** | **91.6** | **75.0** | **100.0** | **91.9** |

Table 9. Spellchecking results on Dataset B. Accuracy in %.

| Case | Total |
| --- | --- |
| Correct | 107 |
| Total | 108 |
| Accuracy | 99.1 |

### 5.2 Failure Cases

In this section, we analyze the erroneous cases of the proposed approach. Table 10 presents some failure cases in the rows. Each row contains a sentence with a misspelled word, a sentence with the correct word, and a sentence with the corrected word. After analyzing these cases, we identified the following causes:

1. Language model: In the first two cases (i.e., the top two rows), the language model assigns a higher likelihood to អាជ្ញាបណ្ណ (meaning *license*) and ស្និទ្ធស្ងាល (meaning *intimate*) over the correct words, អាជ្ញាបត្រ (meaning *license*) and ស្និទ្ធស្នេហ៍ (meaning *intimate*), respectively. This is because អាជ្ញាបណ្ណ and ស្និទ្ធស្ងាល are more commonly used.

2. Multiple correct words: In the third case (i.e., the 3rd row), the proposed approach suggests ពណ៌នា (meaning *describe*) instead of ពំណ៌នា (meaning *describe*) as a correction for ពំណ៌នា. However, both ពណ៌នា and ពំណ៌នា are equally acceptable in the lexicon.

3. Failure to generate enough suggestions: In the last case (i.e., the last row), the proposed approach fails to generate enough suggestions for the language model to rank. បំណឺនជីរីត (misspelled) is a compound of បំណឺន and ជីរីត. Both subwords are misspelled. The proposed approach can suggest the true word (i.e., បំណិន; meaning *skills*) for បំណឺន but not for ជីរីត (true word: ជីវិត; meaning *life*).

---
[9] https://nckl.rac.gov.kh/

Table 10. Failure cases of the proposed approach. Red: misspelled word. Green: true word. Blue: corrected word.

| |
|---|
| យើងមានទំនាក់ទំនងស្និទ្ធស្នេហ៍ជាមួយគ្នា។ |
| យើងមានទំនាក់ទំនងស្និទ្ធស្នេហ៍ជាមួយគ្នា។ |
| យើងមានទំនាក់ទំនងស្និទ្ធស្ងាលជាមួយគ្នា។ |
| ខ្ញុំត្រូវការអាជ្ញបំត្រដើម្បីបើកអាជីវកម្មថ្មី។ |
| ខ្ញុំត្រូវការអាជ្ញាបត្រដើម្បីបើកអាជីវកម្មថ្មី។ |
| ខ្ញុំត្រូវការអាជ្ញាបណ្ណដើម្បីបើកអាជីវកម្មថ្មី។ |
| គាត់បានពំណ៌នាអំពីទេសភាពដ៏ស្រស់ស្ងាត។ |
| គាត់បានពំណ៌នាអំពីទេសភាពដ៏ស្រស់ស្ងាត។ |
| គាត់បានពណ៌នាអំពីទេសភាពដ៏ស្រស់ស្ងាត។ |
| ការអប់រំជួយអភិវឌ្ឍបំណឺនជីរីតសម្រាប់កុមារ។ |
| ការអប់រំជួយអភិវឌ្ឍបំណិនជីវិតសម្រាប់កុមារ។ |
| ការអប់រំជួយអភិវឌ្ឍបំណិនជីរីតសម្រាប់កុមារ។ |

### 5.3 Comparisons with NextSpell and KhmerLang

NextSpell is a commercial Khmer spellchecking tool that can detect and automatically correct spelling mistakes. In NextSpell, there can be four cases:

1. No detection (ND): NextSpell is unable to detect any spelling mistakes, even though they are present.

2. No correction (NC): NextSpell is able to detect spelling mistakes but is unable to correct them.

3. Wrong correction (WC): NextSpell is able to detect spelling mistakes but corrects them wrongly.

4. Right correction (RC): NextSpell is able to detect spelling mistakes and corrects them accurately.

On Dataset A, Table 11 shows that NextSpell achieves an accuracy rate (i.e., RC) of approximately 9.9%, which is significantly lower than that of our proposed approach (i.e., up to 94.4% in Table 8). The table also shows that many misspelled words (i.e., 79.3%) can be detected but not corrected. Additionally, approximately 9.6% of errors are not detected at all.

Table 11. NextSpell results on Dataset A.

| Case | % |
|---|---|
| ND | 9.6 |
| NC | 79.3 |
| WC | 1.2 |
| RC | 9.9 |

On the other hand, KhmerLang is an open-source Khmer spellchecking tool that can detect and suggest possible corrections but does not perform auto-correction. On Dataset B, we compare the proposed approach with KhmerLang and NextSpell. Since Dataset B contains person names that do not exist in the lexicon, a spellchecking tool should not flag those names as misspelled words. Table 12 shows that our proposed approach can identify names and flag them as not misspelled significantly better (i.e., 99.1%) than both NextSpell (i.e.,

46.3%) and KhmerLang (i.e., 32.4%). Compared with KhmerLang, NextSpell performs better with names.

Table 12. NextSpell, KhmerLang, and our results (%) on Dataset B. **Bold**: best.

| Method | Misspelled | Not Misspelled |
|---|---|---|
| NextSpell | 53.7 | 46.3 |
| KhmerLang | 67.6 | 32.4 |
| Ours | **0.9** | **99.1** |

### 5.4 Future Work

Future work should include the following tasks:

1. Balancing the influence of the language model can enhance spellchecking accuracy. Over-reliance on a language model may favor commonly used words while neglecting rare words and compounds.

2. Generating a sufficient number of possible suggestions for the language model to rank can also enhance spellchecking accuracy. This step needs to ensure that true corrections are included among the possible suggestions.

3. Expanding the benchmark datasets will allow for a more thorough evaluation.

4. Improving each stage or component can enhance overall spellchecking accuracy. Since the proposed approach relies on multiple stages, such as word segmentation, G2P, and NER, further training on larger datasets could improve each of these components, and thus the overall accuracy of the spellchecking system.

## 6 Conclusion

We proposed the first holistic approach for Khmer spellchecking, addressing multiple challenges associated with the Khmer language. We incorporated a Khmer language model to introduce contextual information into the spellchecking process. Experimental results demonstrate that our approach achieves the state-of-the-art performance in Khmer spellchecking. Additionally, we have contributed the first open benchmark datasets for Khmer spellchecking and NER tasks.

**Acknowledgment**

We would like to acknowledge the contributions of the intern students, namely Mr. Keat Charavuth and Mr. Yin Sambat.